# STARS: Sparse Learning Correlation Filter with Spatio-temporal Regularization and Super-resolution Reconstruction for Thermal Infrared Target Tracking

Shang Zhang[1,2,3*], Xiaobo Ding[1,2,3], Huanbin Zhang[1,2,3], Ruoyan Xiong[1,2,3], and Yue Zhang[1,2,3]

[1] College of Computer and Information Technology, China Three Gorges University, Hubei, Yichang, 443002, China
[2] Hubei Province Engineering Technology Research Center for Construction Quality Testing Equipment, China Three Gorges University, Yichang 443002, China
[3] Hubei Key Laboratory of Intelligent Vision Based Monitoring for Hydroelectric Engineering, China Three Gorges University, Yichang 443002, China
zhangshang@ctgu.edu.cn

**Abstract.** Thermal infrared (TIR) target tracking methods often adopt the correlation filter (CF) framework due to its computational efficiency. However, the low resolution of TIR images, along with tracking interference, significantly limits the performance of TIR trackers. To address these challenges, we introduce STARS, a novel sparse learning-based CF tracker that incorporates spatio-temporal regularization and super-resolution reconstruction. First, we apply adaptive sparse filtering and temporal domain filtering to extract key features of the target while reducing interference from background clutter and noise. Next, we introduce an edge-preserving sparse regularization method to stabilize target features and prevent excessive blurring. This regularization integrates multiple terms and employs the alternating direction method of multipliers to optimize the solution. Finally, we propose a gradient-enhanced super-resolution method to extract fine-grained TIR target features and improve the resolution of TIR images, addressing performance degradation in tracking caused by low-resolution sequences. To the best of our knowledge, STARS is the first to integrate super-resolution methods within a sparse learning-based CF framework. Extensive experiments on the LSOTB-TIR, PTB-TIR, VOT-TIR2015, and VOT-TIR2017 benchmarks demonstrate that STARS outperforms state-of-the-art trackers in terms of robustness.

## 1  Introduction

Thermal infrared (TIR) target tracking is a critical area of research in computer vision. While RGB-based visual target tracking faces significant challenges under adverse weather conditions, such as fog, low light, and rain, TIR tracking remains effective in these scenarios due to its robustness against illumination changes. This characteristic

*Corresponding author

2         XXX et al.

makes TIR tracking particularly valuable for applications in domains such as driver assistance systems, environmental monitoring, and industrial automation. Numerous efforts have been made to improve the performance of TIR object tracking methods. Among these, the correlation filter (CF) framework has gained popularity due to its computational efficiency. However, the effectiveness of TIR trackers is often limited by challenges such as object deformation, occlusion, noise, and unreliable training samples.

In recent years, CF-based trackers have emerged as some of the most widely used methods for target tracking, owing to their ability to achieve high accuracy while maintaining fast processing speeds. Among these trackers, the discriminative correlation filter (DCF) [1]-[5] has garnered significant attention for its effectiveness. DCF addresses the ridge regression problem in the Fourier frequency domain by replacing traditional ridge regression operations with efficient element-wise multiplication. This approach not only reduces computational complexity but also enhances tracking precision and speed. However, DCF operates under the assumptions of a uniform background and stable target motion. Consequently, it struggles to maintain accurate tracking when the target undergoes deformation or when the background changes dynamically. Given the critical role of thermal infrared target tracking in real-world applications, substantial efforts have been made to overcome these limitations [8], [10]. In this context, Jiri Matas et al. [1] proposed the kernelized correlation filter (KCF) [4], which employs a Gaussian kernel function [4], [7] to map the linear ridge regression problem into a nonlinear space. This transformation effectively addresses the challenge of target deformation. Building upon this, Danelljan et al. [9] introduced the spatially regularized DCF (SRDCF), which incorporates spatial regularization to suppress boundary effects, thereby significantly mitigating the impact of target de formation. Furthermore, the DeepSRDCF method integrates deep convolutional neural networks with spatial regularization, enhancing tracking performance by improving background adaptability and robustness to variations in target appearance.

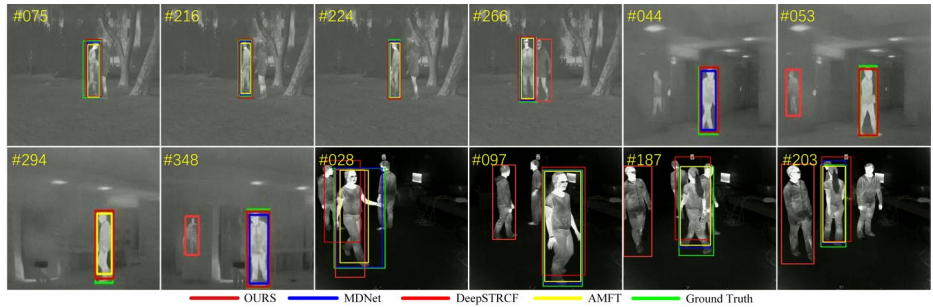

**Fig. 1.** The visualization shows tracking results for four trackers, with blue, rose, yellow, and red boxes representing MDNet, DeepSTRCF, AMFT, and the proposed STARS, respectively.

Challenges such as target blurring, occlusion, and deformation significantly affect the accuracy of TIR tracking methods. While techniques like sparse representation, Siamese networks, and convolutional neural networks have been used to address these issues, they still have certain limitations. To overcome these, we propose STARS, a novel sparse learning-based tracker that integrates spatio-temporal regularization and super-resolution reconstruction. By leveraging the computational efficiency of the CF framework, STARS uses a DCF to combine adaptive sparse representation with spatio-temporal regularization constraints, improving localization accuracy and reducing excessive target blurring. A key innovation of STARS is its gradient-enhanced super-resolution method, specifically designed to address the challenges of low-resolution images. This method extracts fine-grained features from low-resolution images and optimizes them iteratively to reconstruct higher-resolution images. Extensive experimental evaluations show that our STARS outperforms state-of-the-art trackers in terms of robustness.

The main contributions of this work are summarized as follows:

- We introduce a novel objective function that integrates adaptive sparse representation and temporal regularization to reduce the impact of noise while enhancing target localization accuracy.
- We propose a model that combines temporal consistency and sparsity constraints to reduce redundant features during tracking and prevent excessive blurring.
- We present a novel gradient-enhanced super-resolution method to improve the resolution of TIR images, addressing performance degradation in tracking caused by low-resolution sequences.
- We propose a robust perception algorithm for extracting fine-grained object edge features, increasing the tracker's sensitivity to edges and enhancing the reconstruction quality of TIR images.

The structure of this paper is organized as follows: Section II offers an overview of discriminative DCF tracking and fine-grained feature extraction methods. Section III provides a detailed explanation of the proposed STARS and its key components. Section IV presents both qualitative and quantitative evaluations of the proposed STARS, comparing its performance with that of state-of-the-art trackers. Finally, Section V summarizes the key findings and suggests directions for future research.

## 2 Related Work

### 2.1 Tracking with Correlation Filter

In recent years, CF-based trackers have garnered significant attention in the field of visual tracking due to their high computational efficiency. Bolme et al. [12] were the first to introduce CF technology for target tracking with the development of the MOSSE tracker. Their approach transformed the tracking problem into a frequency-domain operation, substantially improving computational efficiency. Despite utilizing a single-channel grayscale feature, the MOSSE tracker demonstrated performance comparable to other state-of-the-art trackers of its time, generating considerable interest within the research community. Building upon this foundation, Henriques et al. [15] proposed the Circulant Structure with Kernels (CSK) tracker, which enhanced robustness by replacing single-channel features with hand-crafted features. This work was further extended to the Kernelized Correlation Filter (KCF), which incorporated multi-channel features such as Histogram of Oriented Gradients (HOG) to capture the spatial appearance of targets more effectively. Danelljan et al. [16] advanced these methods by introducing a scale pyramid representation to address challenges related to target scaling and deformation, culminating in a 3D correlation filter approach.

However, the accuracy of these trackers tends to degrade as tracking speed increases or when targets undergo deformation or occlusion. To address these limitations, we propose STARS, a novel sparse learning-based CF tracker. Unlike previous trackers, our STARS improves tracking robustness in challenging scenarios, particularly in low-resolution environments, by extracting fine-grained features from the target and using super-resolution to reconstruct low-resolution targets into higher resolution. Moreover, our STARS integrates adaptive sparse representation and spatio-temporal regularization to effectively handle issues such as object blurring and deformation.

### 2.2   Tracking with Modified Correlation Filter

The objective function is a critical determinant of the performance of the original CF tracker. MOSSE [12] pioneered the integration of CF frameworks into target tracking by employing a straightforward functional structure with a regularization term. Notably, significant performance improvements can be achieved by directly refining the objective function within the CF framework. For example, the incorporation of spatial regularization has proven highly effective in addressing challenges posed by object occlusion during tracking. Building on this foundation, Dai et al. [13] enhanced the regularization process by employing the dual alternating direction method of multipliers (ADMM). Similarly, Zhang et al. [14] introduced a spatio-temporally non-local regularization method for CFs. Danelljan et al. [16] further advanced the field with their Discriminative Scale Space Tracker (DSST), which incorporated temporal regularization to successfully mitigate object drift, thereby enhancing tracking robustness.

Building on these advancements, the proposed STARS integrates both spatial and temporal regularization terms into the CF framework, further enhancing tracking performance. By combining these regularization techniques, our STARS effectively manages challenges such as target deformation, occlusion, and noise, ensuring more accurate and stable tracking. This tracker is not only effective for TIR tracking but is also versatile enough to be extended to other CF-based trackers, offering a robust and adaptable solution for a wide range of complex tracking scenarios.

### 2.3   Super-resolution Reconstruction with Correlation Filter

In TIR tracking, factors such as challenging environments, rapid target motion, and occlusion often lead to reduced image resolution. This degradation can result in tracking drift or target loss, particularly when the tracker processes low-resolution (LR) images. To address this issue, various methods have been proposed. Common approaches include interpolation methods, such as bilinear and bicubic interpolation [17], as well as image reconstruction methods based on dictionary learning and sparse representation [20]. For instance, Zeyde et al. [19] introduced a sparse coding method for super-resolution using learned dictionaries, where an overcomplete dictionary was applied to each LR patch to extract sparse coefficients, which were subsequently used to reconstruct high-resolution (HR) patches. Similarly, Zhang et al. [20] utilized dictionary learning and sparse coding to achieve super-resolution for remote sensing images, decomposing LR patches to reconstruct their HR counterparts.

Many existing trackers attempt to utilize deep features to build more robust appearance models for TIR targets. However, the core issue of LR in TIR images remains unresolved. Additionally, the lack of sufficient detail and texture in TIR images intensifies challenges such as target blurring and noise, particularly under LR conditions, which significantly impacts both tracking accuracy and robustness. As a result, improving the resolution of TIR images becomes the primary challenge in TIR target tracking, as LR images inherently limit the effectiveness of tracking algorithms. To address this challenge, we propose a novel gradient-enhanced super-resolution method. This method extracts fine-grained features from low-resolution images and iteratively refines them to reconstruct higher-resolution images. The goal is to enhance tracking performance by improving the clarity and detail of the images.

## 3    Methodology

The proposed STARS integrates three distinct objective functions, each targeting specific aspects of TIR image processing. An overview of STARS, along with its objective functions, is presented in Section 3.1. Section 3.2 focuses on methods for preserving image edges, while Section 3.3 offers a detailed explanation of the method for extracting fine-grained features from TIR images and super-resolution reconstruction.

### 3.1    Preview of STARS

In TIR target tracking, the primary objective is to continuously estimate the state of a designated object based on its initial position and size. To improve the performance, especially in tracking pedestrians within complex backgrounds, and to further improve tracking accuracy and robustness, we define a novel objective function as follows：

$$\min_f \left\{ \frac{1}{2} \parallel \mathcal{F}(x)°f - y \parallel^2 + \alpha_1 \parallel f \parallel_1 + \beta \parallel f - f_{t-1} \parallel^2 \right\} \tag{1}$$

To further enhance the stability and consistency of the proposed STARS, time regularization is integrated into the objective function:

$$\min_f \frac{1}{2} \parallel F(x)°f - y \parallel^2 + \alpha_1 \parallel f \parallel_1 + \alpha_2 \parallel \nabla f \parallel_2^2 + \beta_1 \parallel f - f_{t-1} \parallel^2$$
$$+ \beta_2 \parallel F(x_{t-1})°f_{t-1} - y_{t-1} \parallel^2 + \gamma \parallel f \parallel_2^2 \tag{2}$$

The processing steps of the proposed STARS are illustrated in Fig. 2. Fig. 2(a) depicts the process of locking onto the actual target region in TIR images. Fig. 2(b) illus-trates the update mechanism of the edge-preserving sparse regularization module. Finally, Fig. 2(c) presents the iterative process of processing TIR images and reconstructing the corresponding super-resolution images.

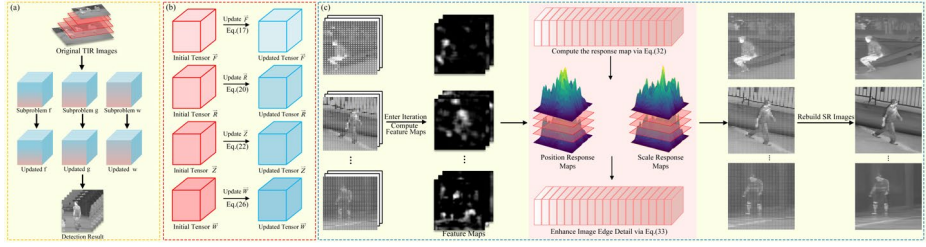

**Fig. 2.** The architecture of the proposed STARS.

**Optimization by the ADMM.** By utilizing the ADMM [21], Eq. (2) can be decomposed into three subproblems, which are then solved iteratively as follows:

$$\begin{cases} f^* = \arg\min_f \frac{1}{2} \parallel F(x)°f - y \parallel^2 + \alpha_1 \parallel f \parallel_1 + \beta_1 \parallel f - f_{t-1} \parallel^2 + \delta_1 \parallel \nabla f \parallel_2^2 \\ g^* = \arg\min_g \frac{1}{2} \parallel F(x_{t-1})°g - y_{t-1} \parallel^2 + \beta_2 \parallel g - f_{t-1} \parallel^2 + \lambda_1 \parallel g \parallel_1 \\ w^* = \arg\min_w \alpha_2 \parallel \nabla f \parallel_2^2 + \gamma \parallel f \parallel_2^2 + \lambda_2 \parallel w - w_r \parallel^2 + \eta \parallel w \parallel_1 \end{cases} \quad (3)$$

The detailed solutions for each subproblem in the iterative process are as follows:

**Subproblem $f$.** To sharpen the focus of the tracker on key features and reduce the influence of irrelevant details, sparse regularization is incorporated into the objective function. The objective for subproblem $f$ is defined as:

$$\mathcal{L}(f) = \frac{1}{2} \parallel y - \mathcal{K}f \parallel_\mathcal{H}^2 + \lambda \frac{\parallel f \parallel_{\mathcal{M}_1}}{\parallel f \parallel_{\mathcal{M}_2} + \epsilon} + \beta \langle f, \mathcal{R}f \rangle_\mathcal{H} \quad (4)$$

where $f$ can be obtained directly through convex optimization. Alternatively, $f$ can also be derived iteratively from Eq. (4) as follows:

$$L(f) = \frac{\parallel y - kf \parallel_H^2}{2} + \lambda \frac{\parallel f \parallel_{M_1}}{\parallel f \parallel_{M_2} + \epsilon} \quad (5)$$

**Subproblem $g$.** The spatial regularization formula is derived from constraints on $\nabla f_k$, $\nabla^2 f_k$, and $\Delta f_k$, ensuring the smoothness and stability of the filter $f$. Therefore, Eq. (6) can be expressed as:

$$\mathcal{R}_{spatial}(f) = a_2 \sum_{k=1}^{L} ((\nabla^2 f_k + \lambda_1 \nabla f_k + \gamma_1 (\nabla f_k)^2 + \mu_1 \nabla^3 f_k)^2 \\ + \lambda_2 (\Delta f_k + \gamma_2 (\Delta f_k)^2 + \mu_2 \nabla^4 f_k)^2 \quad (6)$$

where $f_k^d \in \mathbb{R}$ represents the filter weight corresponding to the $d$-th feature channel at spatial location $k$. The solution is obtained as:

$$\begin{cases} f = \{f_{k,d} \mid k \in \{1,2,\dots,L\}, d \in \{1,2,\dots,D\}\} \\ f_{k,d} = \sum_{i=1}^{H}\sum_{j=1}^{W} w_{i,j,k,d} \end{cases} \quad (7)$$

From Eq. (7), introducing $\alpha$ adjusts the smoothness of the balanced filter. The $p$-norm paradigm enhances control over performance:

$$\alpha_2 = \left\| \sum_{k=1}^{L} (f_k + \lambda_1 \nabla f_k + \gamma_1 (\nabla f_k)^2 + \mu_1 \nabla^3 f_k)^2 \right\|^p \quad (8)$$

Spatial response metrics are introduced to help the tracker capture salient features more effectively:

$$\nabla f_k = \left(\frac{\partial f_k}{\partial x}, \frac{\partial f_k}{\partial y}\right) == \left(\sum_{i=1}^{H-1}\sum_{j=1}^{W}\frac{f_{i+1,j,k}-f_{i,j,k}}{\Delta x}, \sum_{i=1}^{H}\sum_{j=1}^{W-1}\frac{f_{i,j+1,k}-f_{i,j,k}}{\Delta y}\right) \quad (9)$$

By detecting changes in pixel intensity on feature maps and integrating spatial gradient information across all channels, this method mitigates overfitting during the tracking process. The incorporation of second-order gradients enhances the ability to detect subtle variations in target shapes and edges, improving localization accuracy and resilience to spurious fluctuations.

The solution of is obtained as:

$$\nabla^2 f_k = \frac{\partial^2 f_k}{\partial x^2} + \frac{\partial^2 f_k}{\partial y^2} = \frac{\partial^2}{\partial x^2}\left(\sum_{d=1}^{D} f_k^d \cdot \phi_d\right) + \frac{\partial^2}{\partial y^2}\left(\sum_{d=1}^{D} f_k^d \cdot \phi_d\right) \quad (10)$$

**Remark.** Second-order gradients are explicitly incorporated into the detection process. Compared to SRDCF, STARS places a stronger emphasis on second-order gradients in addition to first-order gradients. By capturing intensity curvature variations, STARS effectively suppresses noise and improves precision in pedestrian detection.

**Subproblem $w$.** The temporal regularization formula is defined as:

$$R_{tomporal}(f) = \beta \sum_{m=1}^{P} \left(\delta_t(f_{m,t} - f_{m,t-1} + \epsilon\, sgn(f_{m,t}))\right)^2$$
$$+ \beta_2 \sum_{m=1}^{P} \left(\delta_t^2(f_{m,t} - f_{m,t-1} + \gamma \cdot sgn(f_{m,t-1}))\right)^2 \quad (11)$$

where $sgn(f_{m,t-1})$ represents the direction of the filter from the previous frame, ensuring inter-frame consistency. The parameters $\gamma$ and $\epsilon$ are scaling factors that adjust the influence of each term to maintain smooth transitions in the filter across frames.

To improve model convergence, an additional ADMM solver is applied to subproblem $w$, breaking it down into two additional subproblems, $p$ and $q$, as follows:

$$\begin{cases} \min_f p(f) = \beta_1 \sum_{m=1}^{P} \left(\sqrt{f_{m,t}} - \sqrt{f_{m,t-1}} + \epsilon \cdot sgn(f_{m,t})\right)^2 \\ \min_f q(f) = \beta_2 \sum_{m=1}^{P} \left(k \cdot (f_{m,t} - f_{m,t-1})^2 + \gamma \cdot sgn(f_{m,t-1})\right)^2 \end{cases} \quad (12)$$

**Subproblem $p$.** An offset term $\epsilon \cdot sgn(f_{m,t})$ is introduced, adding a nonlinear component to the solution. The optimal solution to this subproblem is expressed as:

$$f_p^* = \frac{\epsilon}{\beta_1} \cdot sgn(f_{m,t}) + f_{m,t-1} \quad (13)$$

**Subproblem $q$.** A weighting factor $k$ is introduced to increase sensitivity to the magnitude of the changes, resulting in the following solution:

$$f_q^* = \sqrt{\frac{\gamma}{\beta_2}} + f_{m,t-1} \quad (14)$$

### 3.2 Edge Preservation and Sparse Regularization

In this section, we present the Edge Preservation and Sparse Regularization (EPSR) method and propose an alternating optimization algorithm to efficiently learn the components of the tracker. The optimization problem is formulated as follows:

$$\min_{\vec{\mathcal{F}},\vec{\mathcal{R}},\vec{\mathcal{Z}},\vec{\mathcal{W}}} \| \vec{\mathcal{F}} \|_{TNNL} + \lambda_1 \| \vec{\mathcal{Z}} \|_1 + \lambda_2 \| \vec{\mathcal{W}} \|_{(l_1,l_2)} + \lambda_3 \| \vec{\mathcal{R}} - \vec{\mathcal{F}}_{t-1} \|^2 \quad (15)$$

---

**Algorithm 1: Alternating Optimization**

**Input:** $\vec{\mathcal{F}}_{t-1}$ and regularization parameters $\lambda_1, \lambda_2, \lambda_3$.
**Output:** Optimized components: $\vec{\mathcal{F}}, \vec{\mathcal{Z}}, \vec{\mathcal{R}}$ and $\vec{\mathcal{W}}$.
**Initialize:** $\vec{\mathcal{F}}, \vec{\mathcal{Z}}, \vec{\mathcal{R}}, \vec{\mathcal{W}}$ based on $\vec{\mathcal{F}}_{t-1}$, set $k$=0.
**While** not convergence **do**
　Update $\vec{\mathcal{F}}$ by minimizing $\| \vec{\mathcal{F}} \|_{TNNL} + \mu \| \vec{\mathcal{R}} - \vec{\mathcal{F}}_{t-1} \|^2$
　Update $\vec{\mathcal{Z}}$ by minimizing $\lambda_1 \| \vec{\mathcal{Z}} \|_1$
　Update $\vec{\mathcal{W}}$ by minimizing $\lambda_2 \| \vec{\mathcal{W}} \|_{(l_1,l_2)}$
　Update $\vec{\mathcal{R}}$ by minimizing $\lambda_3 \| \vec{\mathcal{R}} - \vec{\mathcal{F}}_{t-1} \|^2$
　$k$=$k$ +1;
**End while**
**Output:** Optimized $\vec{\mathcal{F}}, \vec{\mathcal{Z}}, \vec{\mathcal{R}}$ and $\vec{\mathcal{W}}$ for the current frame

---

The transformation of Eq. (15) into an unconstrained Lagrangian function is expressed as:

$$L(\vec{F},\vec{R},\vec{Z},\vec{W}) = \| \vec{F} \|_{TNNL} + \lambda_1 \| \vec{Z} \|_1 + \lambda_2 \| \vec{W} \|_{(l_1,l_2)} + \lambda_3 \| \vec{R} \|_F^2 + \langle Y_1, \vec{Z} - \vec{F} \rangle$$
$$+ \langle Y_2, \vec{W} - \vec{R} \rangle + \langle Y_3, \vec{R} - \vec{F} \rangle + + \mu \| \vec{Z} - \vec{F} \|_F^2 + \| \vec{W} - \vec{R} \|_F^2 + \| \vec{R} - \vec{F} \|_F^2 \quad (16)$$

where $\mu$ is a positive penalty parameter, and $\mathcal{Y}_i (i=1, 2, 3)$ represents the Lagrange multipliers. Utilizing the ADMM framework, we decompose Eq. (16) into five subproblems, which can be solved by alternately updating the variables.

**Updating $\vec{\mathcal{F}}$.** The update for $\vec{\mathcal{F}}$ in the alternating optimization process is derived as:

$$\vec{\mathcal{F}}^{(k+1)} = \arg\min \left( \| F \|_{TNNL} + \langle Y_1^k, Z - F \rangle + \langle Y_3^k, R - F \rangle + \frac{\mu^k}{2}(\| Z - F \|_F^2 + \| R - F \|_F^2) \right)$$
$$= \arg\min \left\{ \| \vec{\mathcal{F}} \|_{TNNL} + \frac{\mu^k}{2} \left\| \vec{\mathcal{F}} - \left( \vec{\mathcal{Z}} + \frac{Y_1^k}{\mu^k} + \vec{\mathcal{R}} + \frac{Y_3^k}{\mu^k} \right) \right\|_F^2 \right\} \quad (17)$$

Rearranging Eq. (17), we obtain:

$$\| \vec{F} \|_{TNNL} + \frac{\mu^k}{2} \| \vec{F} - \left( \vec{Z} + \vec{R} + \frac{Y_1^k + Y_3^k}{\mu^k} \right) \|^2 \quad (18)$$

This is solved using the tensor singular value thresholding (T-SVT) operator, $D_\tau(\cdot)$, resulting in the update rule for $\vec{F}$:

$$\vec{F}^{(k+1)} = D_{\tau^k} \left( \vec{Z} + \vec{R} + \frac{Y_1^k + Y_3^k}{\mu^k} \right) \quad (19)$$

**Updating $\vec{\mathcal{R}}$.** The update for $\vec{\mathcal{R}}$ is derived as:

$$\vec{\mathcal{R}}^{(k+1)} = \arg\min_{\vec{R}} \lambda_3 \| \vec{R} \|_F^2 + \langle Y_2^k, \vec{W} - \vec{R} \rangle + \frac{\mu^k}{2} \| \vec{R} - \vec{F} \|_F^2$$
$$= \arg\min_{\vec{R}} \lambda_3 \| \vec{R} \|_F^2 + \frac{\mu^k}{2} \left\| \vec{R} - \left( \vec{F} + \vec{W} + \frac{Y_2^k}{\mu^k} \right) \right\|_F^2 \quad (20)$$

This equation can be solved using least squares and regularization, resulting in the update rule:

$$\vec{R}^{(k+1)} = \frac{1}{\lambda_3 + \mu^k} \left( \mu^k \left( \vec{F} + \vec{W} + \frac{Y_2^k}{\mu^k} \right) \right) \quad (21)$$

**Updating $\vec{\mathcal{Z}}$.** The update for $\vec{\mathcal{Z}}$ is derived as:

$$\vec{Z}^{(k+1)} = \arg\min_{\vec{Z}} \left( \lambda_1 \| \vec{Z} \|_1 + \langle Y_1^k, \vec{Z} - \vec{F} \rangle + \frac{\mu^k}{2} \| \vec{Z} - \vec{F} \|_F^2 \right) \quad (22)$$

This subproblem is difficult to optimize directly. The optimization is solved using gradient descent. The gradient of Eq. (22) is given by:

$$\nabla L(\vec{Z}) = \lambda_1 \cdot sgn(\vec{Z}) + \mu^k (\vec{Z} - \vec{F}) \quad (23)$$

To find the optimal solution, we set $\nabla L(\vec{Z}) = 0$:

$$\vec{Z} = \vec{F} - \frac{\lambda_1}{\mu^k} \cdot sgn(\vec{Z}) \quad (24)$$

Eq. (24) is solved using hard thresholding, where $\mu^k$ is a penalty parameter that ensures $\vec{\mathcal{Z}}$ does not deviate excessively from $\vec{\mathcal{F}}$ during the optimization process.

The update rule for $\vec{Z}$ is:

$$\vec{Z}^{(k+1)} = H_{\frac{\lambda_1}{\mu^k}}\left(\vec{F} + \frac{Y_1^k}{\mu^k}\right) \tag{25}$$

**Updating $\vec{\mathcal{W}}$.** The update for $\vec{\mathcal{W}}$ is derived as:

$$\vec{W}^{(k+1)} = \arg\min_{\vec{W}} \lambda_2 \parallel \vec{W} \parallel_{(l_1,l_2)} + \frac{\mu^k}{2}\left\|\vec{W} - \left(\vec{R} + \frac{Y_2^k}{\mu^k}\right)\right\|_F^2 \tag{26}$$

To solve Eq. (26) using the Iterative Shrinkage Thresholding Algorithm (ISTA), the gradient of the objective function is first calculated:

$$\mu^k(\vec{W} - \vec{R}^{(k)}) + Y_2^k \tag{27}$$

Then, the proximal operator for the regularization term $\lambda_2 \parallel \vec{W} \parallel_{(l_1,l_2)}$, denoted as $prox_{\lambda_2}(\vec{V})$, is then applied. The proximal operator can be expressed as:

$$prox_{\lambda_2}(\vec{V}) = S_{\lambda_2}(\vec{V}) = sign(\vec{V}) \cdot max\{|\vec{V}| - \lambda_2, 0\} \tag{28}$$

where $\vec{V}$ is the input vector for the soft-thresholding operation. Specifically, in this context:

$$\vec{V} = \vec{R}^{(k)} + \frac{Y_2^k}{\mu^k} - \frac{1}{\mu^k}Y_2^k \tag{29}$$

The solution of Eq. (29) is:

$$\vec{W}^{(k+1)} = \vec{R}^{(k)} - \frac{1}{\mu^k}\nabla\left(\lambda_2 \parallel \vec{W} \parallel_{(l_1,l_2)}\right) = S_{\frac{\lambda_2}{\mu^k}}\left(\vec{R}^{(k)} + \frac{Y_2^k}{\mu^k}\right) \tag{30}$$

**Updating Lagrangian Multipliers $\mathcal{Y}_1, \mathcal{Y}_2, \mathcal{Y}_3$ and Penalty Parameter μ.** The updates for the Lagrangian multipliers $\mathcal{Y}_1, \mathcal{Y}_2, \mathcal{Y}_3$ and the penalty parameter $\mu$ are as follows:

$$\begin{cases} \mathcal{Y}_1^{(k+1)} = \mathcal{Y}_1^k + \mu^k(\vec{Z}^{(k+1)} - \vec{\mathcal{F}}^{(k+1)}) \\ \mathcal{Y}_2^{(k+1)} = \mathcal{Y}_2^k + \mu^k(\vec{\mathcal{W}}^{(k+1)} - \vec{\mathcal{R}}^{(k+1)}) \\ \mathcal{Y}_3^{(k+1)} = \mathcal{Y}_3^k + \mu^k(\vec{\mathcal{R}}^{(k+1)} - \vec{\mathcal{F}}^{(k+1)}) \\ \mu^{(k+1)} = \rho * \mu^k \end{cases} \tag{31}$$

where, $\mathcal{Y}_1, \mathcal{Y}_2, \mathcal{Y}_3$ are the Lagrangian multipliers associated with the constraints in the optimization problem. $\mu$ is the penalty parameter, controlling the trade-off between the objective function and the constraints. While $\rho$ is a constant scaling factor that is typically set to slightly increase $\mu$ after each iteration, ensuring convergence of the optimization process.

---

**Algorithm 2: Updates of the EPSR Model**

**Input:** $\vec{\mathcal{F}}_{t-1}, \lambda_1, \lambda_2, \lambda_3$
**Output:** $\vec{\mathcal{F}}, \vec{\mathcal{Z}}, \vec{\mathcal{R}}, \vec{\mathcal{W}}$

**Initialize:** $\vec{\mathcal{F}}, \vec{\mathcal{Z}}, \vec{\mathcal{R}}, \vec{\mathcal{W}}$ based on $\vec{\mathcal{F}}_{t-1}$, $\mathcal{Y}_1^0 = \mathcal{Y}_2^0 = \mathcal{Y}_3^0 = 0$, $k=0$
**while** not convergence **do**
  Update $\vec{F}^{(k+1)}$ using Eq. (19)
  Update $\vec{R}^{(k+1)}$ using Eq. (21)
  Update $\vec{Z}^{(k+1)}$ using Eq. (25)
  Update $\vec{W}^{(k+1)}$ using using Eq. (29)
  Update $\mathcal{Y}_1^{(k+1)}$, $\mathcal{Y}_2^{(k+1)}$ and $\mathcal{Y}_3^{(k+1)}$ using Eq. (30)
  Update $\mu^{(k+1)}$ using Eq. (31)
  Check convergence condition
  $k = k + 1$
**end while**
**Output:** Optimized $\vec{\mathcal{F}}, \vec{\mathcal{Z}}, \vec{\mathcal{R}}$ and $\vec{\mathcal{W}}$ for the current frame

### 3.3 Super-resolution Reconstruction with Fine-grained Features

In TIR tracking, especially when dealing with low-resolution or blurred target images, tracking accuracy can suffer significantly degrade. To overcome this challenge, we propose a novel gradient-enhanced super-resolution (GESR) method, which improves the resolution of low-quality TIR images by emphasizing fine-grained features. This method is integrated into our STARS tracker to enhance its robustness, particularly in environments where low-resolution images are common.

To improve computational efficiency, we employed element-wise multiplication, and the operations are carried out in the frequency domain, as expressed by:

$$\hat{\mathcal{Y}} = \sum_{d=1}^{\mathcal{D}} \hat{\mathcal{W}}_k \odot \hat{\mathcal{D}}_k \tag{32}$$

where $\hat{\mathcal{Y}}$ is the Fourier transform of the response map. $\hat{\mathcal{W}}_k$ denotes the weight map for the feature at the $k$-th position. $\hat{\mathcal{D}}_k$ is the frequency-domain representation of the target feature. $\odot$ represents element-wise multiplication in the frequency domain.

Once the response map is computed, the target position is identified by locating the maximum value within the map. Fine-grained features are then extracted from the target region and used to reconstruct high-resolution images. This process is aided by gradient analysis in the local region surrounding the target, which helps enhance the detection of target details. To mitigate background clutter and other unwanted influences, sparse regularization is introduced, facilitating the extraction of fine-grained features from the target region. This approach significantly improves the tracking accuracy of our STARS tracker. The objective function for this algorithm is formulated as follows:

$$\mathcal{L} = \sum_{i=1}^{N} \left( m \parallel X \parallel_1 + q \parallel \hat{\mathcal{Y}} \odot \nabla \mathcal{X} - \nabla \mathcal{W} \parallel_2^2 \right) \tag{33}$$

where $\nabla \mathcal{X}$ denotes the gradient of the target image, and $\nabla \mathcal{W}$ denotes the gradient of the reference image. By applying element-wise multiplication, gradient consistency is promoted, further improving the robustness of the feature extraction process. Inspired by the image reconstruction method proposed in [23], we introduce a super-resolution reconstruction objective function based on adaptive sparse representation and CF. We employ an iterative optimization approach to solve the super-resolution reconstruction problem, with the objective function expressed as:

$$L_{SR} = \mathcal{L} + \sum_{i=1}^{N} \left(\lambda_1 \parallel X - R(I_{LR}) \parallel_2^2 + \lambda_2 \parallel I_{HR} - \hat{I}_{HR} \parallel_2^2\right) + \eta \parallel \hat{I}_{HR} * h - I_{HR} \parallel_2^2 \quad (34)$$

At this stage, we already have a preliminary estimate of the near-super-resolution image while extracting the fine-grained features, which helps to reduce errors before optimization begins. Eq. (33) enhances the edge details of the image, contributing to better super-resolution reconstruction in later stages. In each iteration, we up-sample the low-resolution (LR) image and measure the difference from the original LR image. In the image frequency domain, CF is used to enhance edges and details, while the coefficient $\eta$ is gradually adjusted based on previous iterations. This coarse-to-fine iterative process continues until the reconstructed image meets the detail requirements or the maximum iteration number $T_{max}$ is reached.

We stop the iteration when the change in pixel values between iterations is smaller than a predefined threshold, as measured by:

$$\parallel \hat{I}_t - \hat{I}_{t-1} \parallel_2^2 < \epsilon \quad (35)$$

## 4    Experiment

In this section, we evaluate the performance of STARS across four benchmarks include LSOTB-TIR [24], PTB-TIR [25], VOT-TIR 2015 [26], and VOT-TIR 2017 [27]. First, we outline the experimental setup, including the evaluation metrics and the benchmarks used. Next, we explore the robustness of the algorithm and identify the optimal values of key model parameters through ablation experiments. Finally, STARS is compared with several state-of-the-art trackers, and the tracking results are visualized for analysis.

### 4.1    Implementation Details

**Experimental Platform and Evaluation Criteria.** To ensure a thorough and objective evaluation, we follow the protocols outlined in [24] and [25], using the OPE method to assess tracker performance. For the PTB-TIR dataset, both the success rate and precision metrics are employed. The success rate is calculated by measuring the Intersection over Union (IoU) between the ground truth and predicted bounding boxes. Precision is determined by averaging the IoU values across all frames, providing an overall accuracy measure of the tracker. Additionally, for the LSOTB-TIR dataset, we introduce Normalized Precision (NP) to offer a more comprehensive comparison of tracker performance. NP is calculated by averaging the ratio of IoU to the true bounding box area across all frames, accounting for variations in object appearance and size, as well as dynamic changes in the scene, thus offering a more reliable assessment. The proposed STARS is implemented using Python 3.7.1, and all experiments are conducted on a system running the Ubuntu 20.04 operating system, with an Intel i7-12700H CPU@2.30GHz and an NVIDIA RTX 4060 Ti GPU.

## 4.2 Ablation Experiment

**Ablation Study of Methods.** We perform experiments to evaluate the effectiveness of each component of the proposed STARS. First, we compare the KCF-ASTF model with the baseline KCF model to demonstrate the stability improvements achieved by integrating adaptive sparsity and the temporal filter. Next, we evaluate the KCF-EPSR model against the baseline to highlight the contribution of sparse regularization. Finally, we compare the KCF-GESR model with KCF to showcase the effectiveness of the proposed gradient-enhanced super-resolution method.

**Table 1.** Ablation Experiment on PTB-TIR Benchmark

| Model | Component | | | Precision (P/%) | Success (S/%) | Speed (FPS) |
|---|---|---|---|---|---|---|
| | ASTF | EPSR | GESR | | | |
| KCF(baseline) | | | | 59.0 | 40.0 | **301** |
| KCF-ASTF | √ | | | 74.1 | 49.2 | 32.2 |
| KCF-EPSR | | √ | | 72.6 | 53.5 | 40.3 |
| KCF-GESR | | | √ | 81.2 | 56.8 | 15.9 |
| STARS | √ | √ | √ | **83.0** | **62.8** | 12.1 |

The results of the ablation study on PTB-TIR benchmark are presented in Table 1, with KCF serving as the baseline model. First, KCF-ASTF outperforms KCF by 15.1% in accuracy and 9.2% in success rate, demonstrating the positive impact of adaptive sparsity and temporal filters in enhancing the stability of the tracker during operation. Next, incorporating EPSR into the baseline model results in improvements of 13.6% in accuracy and 13.5% in success rate. EPSR significantly improves tracking accuracy, primarily by reducing image blur caused by rapid target motion, thereby boosting overall performance. When compared to KCF-ASTF, the addition of sparse regularization, which introduces a level of sparsity to the module, notably enhances tracking speed. Finally, KCF-GESR shows a substantial increase in performance, with a 22.2% improvement in accuracy and a 16.8% gain in success rate over KCF. The gradient-enhanced super-resolution method (GESR) enhances the ability of the tracker to accurately identify targets, especially in complex scenarios. In conclusion, each component of the tracker plays an essential role in boosting tracking performance, emphasizing the importance of ASTF, EPSR, and GESR in improving both accuracy and robustness.

**Ablation Study of Parameter.** To further investigate the impact of the regularization term coefficient $m$ on the performance of STARS, Fig. 3 presents the accuracy and success rate of the tracker on the PTB-TIR dataset for different values of $m$.

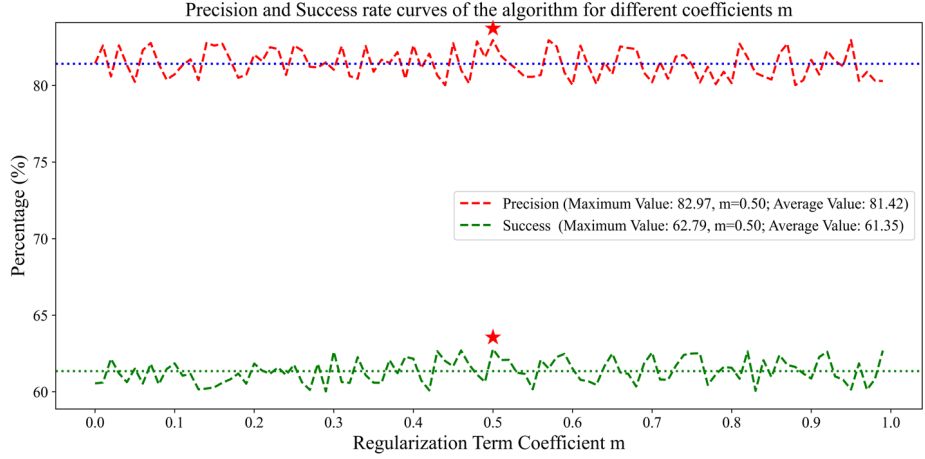

**Fig. 3.** Precision and success rate of the tracker with different regularization coefficient $m$.

The results show that when $m$ is set to 0.50, the tracker achieves the highest precision (82.97%) and success rate (62.79%). Therefore, in this study, $m$ is set to 0.50 to balance preserving the integrity of the target features and enhancing the extraction of fine-grained target characteristics.

### 4.3    Performance Comparison with State-of-the-arts

**Results on PTB-TIR.** The comparison experiments between our STARS tracker and several other trackers, including KCF [4], SRDCF [9], DeepSTRCF [28], ECO-HC [29], Staple [30], and DSST [31], on the PTB-TIR benchmark are presented in Fig. 4. As shown in Fig. 4(a) and (b), the STARS tracker outperforms all the others, achieving the highest accuracy (83.0%) and success rate (62.8%) on the PTB-TIR dataset. When compared to the SRDCF tracker, which is based on correlation filters, STARS improves both accuracy and success rate by 2.8% and 4.0%, respectively. The superior performance of STARS on the PTB-TIR dataset can be attributed to the innovative adaptive sparsity and temporal filter modules. The adaptive sparse representation guides the tracker to initially focus on the general region of the target, and then iteratively refines the target image. This iterative process significantly enhances tracking precision by progressively improving the accuracy of target localization.

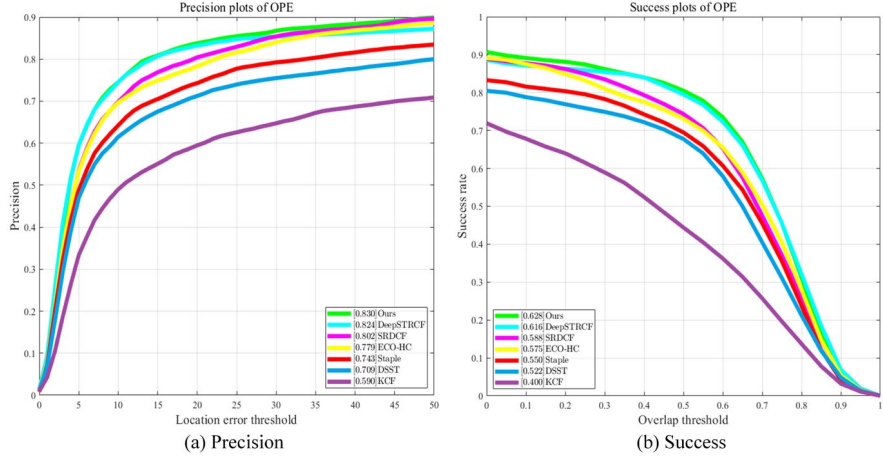

**Fig. 4.** Performance comparison results on PTB-TIR benchmark.

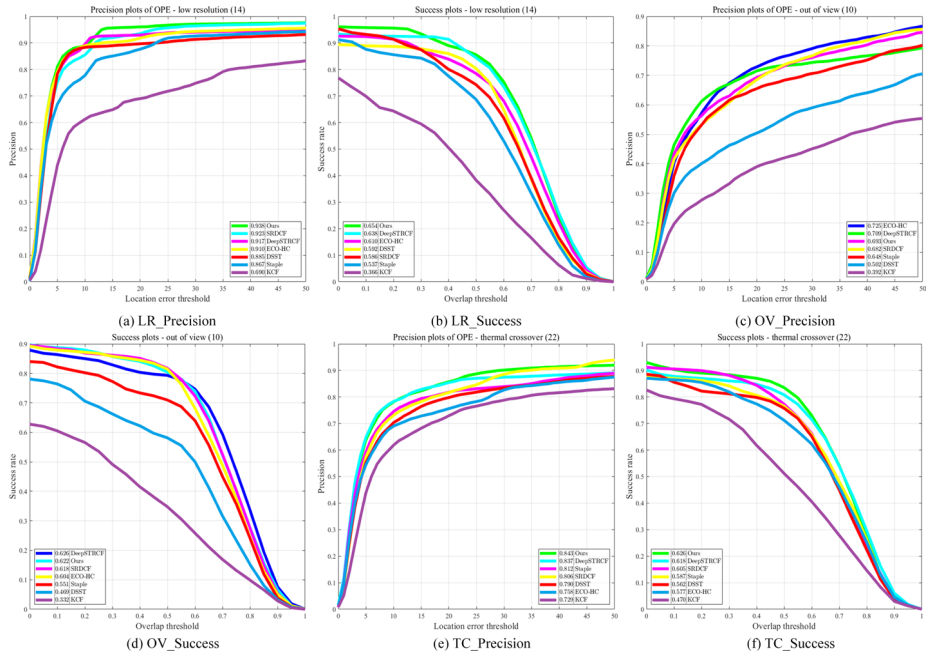

**Fig. 5.** Performance comparison results on the PTB-TIR benchmark in three scenarios (scenarios including low resolution, out of view, thermal crossover).

Comparison experiments of our tracker with DeepSTRCF, SRDCF, ECO-HC, Staple, DSST, and KCF across three challenging scenarios in the PTB-TIR benchmark—low resolution, out of view, and thermal crossover—are shown in Fig. 5. As depicted in Fig. 5, our proposed tracker achieves the highest precision and success rate in both

16     XXX et al.

low-resolution and thermal-crossover scenarios. The excellent performance of STARS in low-resolution scenes is attributed to the proposed fine-grained edge feature detection algorithm. By extracting fine-grained features from the target image and applying stepwise iterative optimization, the tracker is guided to progressively focus on enhancing key fine-grained features of the low-resolution image, significantly improving both precision and success rate.

**Results on LSOTB-TIR.** To further validate the effectiveness of our STARS tracker on challenging attributes, we conducted experiments with various trackers across 12 attributes and 4 scenarios defined by the LSOTB-TIR benchmark, as shown in Table 2. Since our proposed tracker is built upon the baseline model, its performance is influenced by the tracking algorithm and features it incorporates.

As shown in Table 2, STARS achieves the best overall accuracy (74.3%), normalized accuracy (66.2%), and success rate (58.5%) on the LSOTB-TIR dataset. It outperforms other trackers in 3 out of the 12 attributes, achieving the highest accuracy in 3 attributes and 3 scenarios, and the best success rate in 3 attributes. While ECO-HC achieves the highest accuracy on the FM attribute, its overall performance lags behind STARS by 2.6, 1.5, and 1.5 percentage points in accuracy, normalized accuracy, and success rate, respectively.

**Table 2.** Tracking Challenges Results on LSOTB-TIR Benchmark.

| Attributes Type | Attributes Name | ECO-HC | DeepSTRCF | SRDCF | OURS |
|---|---|---|---|---|---|
| Challenge | Deformation | **75.6**/66.5/61.2 | 72.4/62.6/56.2 | 71.3/62.5/52.7 | 72.1/**67.2**/**62.3** |
| | Occlusion | 71.9/**67.2**/**60.7** | 65.7/58.5/50.5 | 69.2/62.3/53.2 | **73.4**/65.0/59.6 |
| | Scale Variation | 68.5/61.4/53.4 | 70.8/59.2/52.4 | **74.6**/**67.0**/**59.5** | 70.6/62.8/56.0 |
| | Background Clutter | 72.5/67.4/61.0 | 73.4/67.1/60.4 | 71.0/66.7/60.4 | **74.8**/**69.3**/**61.2** |
| | Low Resolution | 81.2/63.8/45.8 | 82.4/68.6/56.7 | 80.2/64.0/57.3 | **84.2**/**71.7**/**62.8** |
| | Fast Motion | **80.7**/**74.3**/62.4 | 75.8/72.0/**63.6** | 69.4/61.8/53.9 | 71.1/64.2/56.8 |
| | Motion Blur | 72.6/66.0/57.3 | **75.6**/**69.8**/**62.1** | 73.6/63.2/55.8 | 72.9/64.5/55.4 |
| | Out of View | 70.6/66.2/**61.2** | 74.3/**70.2**/60.6 | 72.2/68.4/60.2 | **76.6**/67.3/57.1 |
| | Intensity Variation | 69.3/64.8/53.1 | **78.6**/**74.8**/**64.3** | 74.5/66.3/54.8 | 75.0/68.6/58.2 |
| | Thermal Crossover | 59.2/49.7/48.5 | 59.4/51.8/49.6 | 63.2/**56.6**/48.4 | **65.6**/54.0/**50.2** |
| | Aspect Ratio Variation | 67.0/59.2/53.3 | 73.4/**69.2**/51.6 | **76.2**/67.6/**58.3** | 72.9/64.2/56.4 |
| | Distractor | 63.6/54.0/50.7 | 66.4/60.8/49.2 | 70.3/61.6/55.4 | **72.8**/**64.5**/**57.2** |
| Scenario | Vehicle-mounted | 79.5/72.6/67.2 | 81.2/73.8/**68.4** | 82.3/74.6/66.8 | **83.6**/**75.3**/67.0 |
| | Drone-mounted | 65.7/62.7/53.6 | 71.2/67.4/58.6 | 66.2/60.0/52.6 | **73.6**/66.2/56.8 |
| | Surveillance | 76.1/**73.0**/**65.8** | 62.9/56.0/51.2 | 71.2/65.4/58.6 | **79.5**/72.6/64.2 |
| | Hand-held | 73.5/66.1/56.2 | **77.8**/**68.6**/**60.2** | 67.8/57.4/50.4 | 70.3/62.8/55.4 |
| Challenge and Scenario | All | 71.7/64.7/57.0 | 72.6/65.7/57.2 | 72.1/64.1/56.1 | **74.3**/**66.2**/**58.5** |

As shown in Table 3, the LSOTB-TIR benchmark is widely used for evaluating tracking methods, and we collected and analyzed the experimental results of 12 trackers on this benchmark. The top three methods in terms of accuracy scores are STARS, TransT, and ECO-TIR. Among these, Transformer-based TransT demonstrates exceptional performance in complex scenes. In contrast, the proposed STARS and ECO-TIR, both CF-based trackers, show significant potential in challenging TIR scenarios due to their strong target recognition capabilities. Compared to the CF-based ECO-TIR tracker, STARS improves tracking accuracy rate and success rate by 3.5% and 4.5%, respectively.

Notably, STARS ranks first in precision, thanks to the combination of the proposed adaptive sparse and temporal filter with the fine-grained edge feature detection algorithm.

**Table 3.** Comparative Results on LSOTB-TIR Benchmark.

| Method | Tracker | LSOTB-TIR | | | Speed (FPS) |
|---|---|---|---|---|---|
| | | Pre. ↑ | Norm. Pre. ↑ | Suc. ↑ | |
| Siamese Network | SiamSAV | 0.700 | - | 0.580 | 16 |
| | CFNet-TIR | 0.580 | 0.540 | 0.478 | 29 |
| Transformer | ASTMT | 0.711 | - | 0.657 | 45 |
| | TransT | 0.798 | - | 0.673 | 44 |
| Deep learning | SuperDiMP | 0.764 | 0.692 | 0.647 | 28 |
| | AMFT | 0.765 | - | 0.616 | 7 |
| | HSSNet | 0.515 | - | 0.409 | 18 |
| | MLSSNet | 0.596 | 0.549 | 0.459 | 27 |
| | MDNet | 0.750 | 0.686 | 0.601 | 2 |
| Correlation Filter | ECO-TIR | 0.768 | 0.695 | 0.630 | 20 |
| | MCFTS | 0.640 | - | 0.480 | 4 |
| | ATOM | 0.729 | 0.647 | 0.595 | 34 |
| | STARS | **0.803** | **0.713** | **0.675** | 11.3 |

**Results on VOT-TIR 2015 and 2017.** We evaluated the proposed STARS tracker against state-of-the-art methods using the VOT-TIR2015 and VOT-TIR2017 benchmarks. Performance metrics are in Table IV. STARS achieves the highest accuracy of 0.83 on VOT-TIR2015 and 0.78 on VOT-TIR2017, surpassing SiamRPN, SRDCF, and ECO-MM. It also attains the highest EAO score of 0.342 on VOT-TIR2015 and a competitive 0.296 on VOT-TIR2017. These results demonstrate that the STARS tracker excels in both accuracy and overall performance, showcasing strong robustness and adaptability.

**Table 4.** Performance Comparison on VOT-TIR 2015 and 2017 benchmarks

| Method | Trackers | VOT-TIR 2015 | | | VOT-TIR 2017 | | | Speed |
|---|---|---|---|---|---|---|---|---|
| | | EAO↑ | Acc.↑ | Rob.↓ | EAO↑ | Acc.↑ | Rob.↓ | FPS |
| Deep learning | DeepSTRCF | 0.257 | 0.63 | 2.93 | 0.262 | 0.62 | 3.32 | 6 |
| | ATOM | 0.331 | 0.65 | 2.24 | 0.290 | 0.61 | 2.43 | 30 |
| | Ocean | 0.339 | 0.70 | 2.43 | **0.320** | 0.68 | 2.83 | 25 |
| Siamese network | SiamRPN | 0.267 | 0.63 | 2.53 | 0.242 | 0.60 | 3.19 | 160 |
| | MMNet | 0.340 | 0.61 | 2.09 | 0.320 | 0.58 | 2.90 | 19 |
| | MLSSNet | 0.329 | 0.57 | 2.42 | 0.286 | 0.56 | 3.11 | 18 |
| Transformer | TransT | 0.287 | 0.77 | 2.75 | 0.290 | 0.62 | 3.32 | 6 |
| | DFG | 0.329 | 0.78 | 2.41 | 0.304 | 0.74 | 2.63 | - |
| | CorrFormer | 0.269 | 0.71 | 0.56 | 0.262 | 0.66 | **1.23** | - |
| Correlation filter | SRDCF | 0.225 | 0.62 | 3.06 | 0.197 | 0.59 | 3.84 | 12 |
| | ECO-deep | 0.286 | 0.64 | 2.36 | 0.267 | 0.61 | 2.73 | 16 |
| | MCCT | 0.250 | 0.67 | 3.34 | 0.270 | 0.53 | 1.76 | - |
| | HDT | 0.188 | 0.53 | **5.22** | 0.196 | 0.51 | 4.93 | - |
| | ECO-MM | 0.303 | 0.72 | 2.44 | 0.291 | 0.61 | 2.31 | 8 |
| | STARS(Ours) | **0.342** | **0.83** | 2.40 | 0.296 | **0.78** | 2.68 | 12.8 |

### 4.4     Qualitative Analysis

We conduct tracking experiments on the LSOTB-TIR and PTB-TIR benchmarks using the proposed STARS, MDNet [32], DeepSTRCF [31], and AMFT [33].The visual comparison results for five of the sequences are shown in Fig. 6.

As shown in Fig. 6 (a) and Fig. 6 (c), DeepSTRCF is highly sensitive to target occlusion. When the pedestrian target is partially obstructed by interfering objects, DeepSTRCF often encounters incorrect tracking or tracking failure. In contrast, our STARS utilizes the fine-grained feature extraction algorithm, which allows it to maintain accuracy even in such scenarios. As seen in Fig. 6(e), our tracker successfully tracks the target, even in low-resolution sequences. In comparison, MDNet and DeepSTRCF tend to drift toward similar targets or suffer from tracking loss, resulting in failure to track the correct target.

As shown in Fig. 7, the low resolution of the original TIR image leads to the loss and blurring of pedestrian target details, causing confusion and loss of the target during tracking. In contrast, the proposed gradient-enhanced fine-grained feature extraction and reconstruction significantly enhances the target's resolution, making the edges clearer and improving the overall detail. Experimental results demonstrate that the proposed method outperforms other approaches in terms of tracking performance.

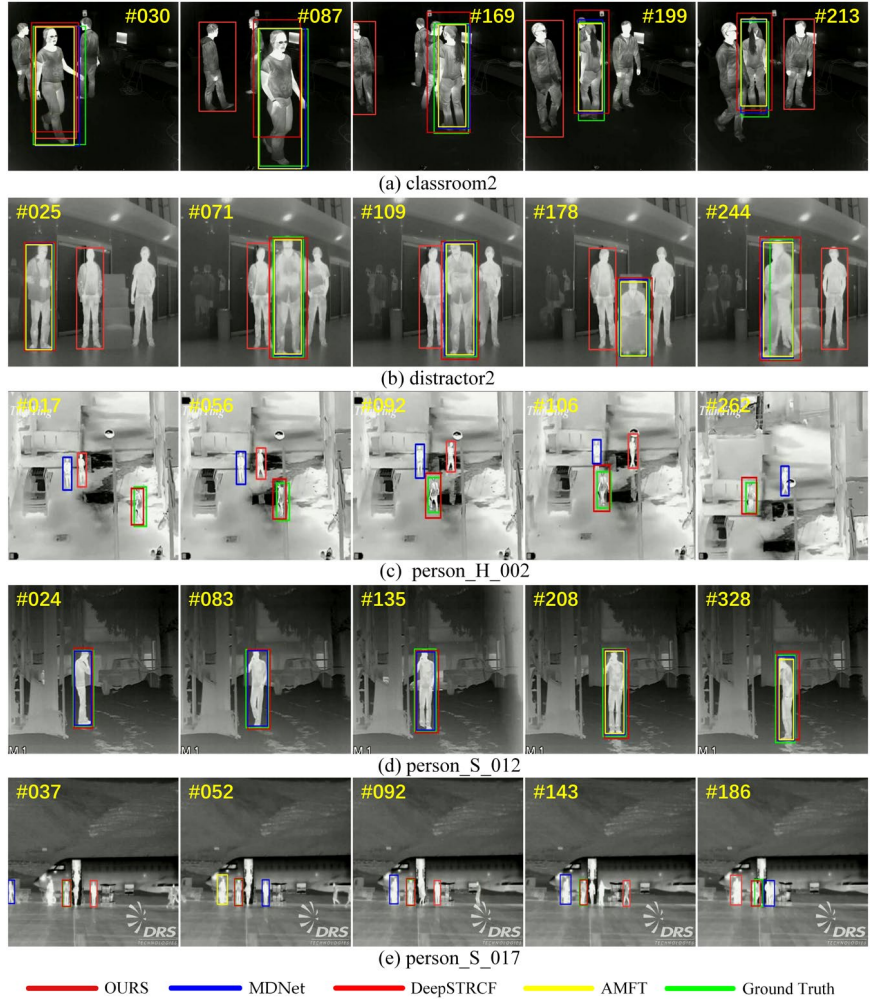

**Fig. 6.** Visual tracking is performed on five challenging sequences from the LSOTB-TIR and PTB-TIR benchmarks using the proposed STARS tracker and three state-of-the-art trackers. (From top to bottom are classroom2, distractor2, person_H_002, person_S_012, person_S_017)

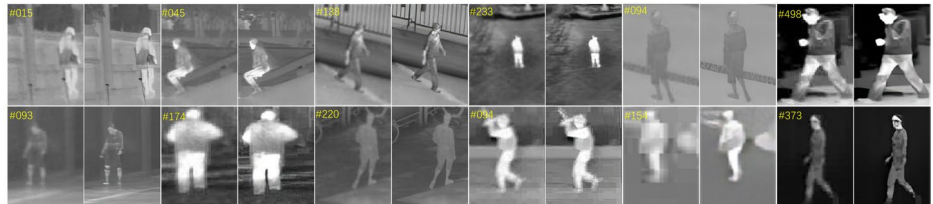

**Fig. 7.** Visual comparison of the enhanced image with the original TIR image by the gradient-enhanced fine-grained feature extraction and reconstruction. (The left-side image represents the original version and is sourced from TIR video sequences in the PTB-TIR and LSOTB-TIR)

## 5      Conclusions

To address the issues of low image resolution and blurring, which severely limit tracker performance in TIR tracking, we propose STARS, a novel correlation filter tracker. By incorporating adaptive sparse regularization and temporal regularization into STARS, we effectively mitigate the impact of background clutter and noise during the training process. Additionally, the integration of edge-preserving and sparse regularization method helps stabilize target features during training. Furthermore, we propose a gradient-enhanced super-resolution method to extract detailed features from TIR images, improving their resolution and, consequently, enhancing tracking accuracy. Extensive experiments on the VOT-TIR2015, VOT-TIR2017, PTB-TIR, and LSOTB-TIR benchmark datasets demonstrate that STARS outperforms other state-of-the-art trackers in terms of robustness.

STARS: Sparse Learning Correlation Filter for Thermal Infrared Target Tracking     21